\documentclass[letterpaper, 10 pt, conference]{ieeeconf}  

\IEEEoverridecommandlockouts                              

\usepackage{graphics} 
\usepackage{epsfig} 
\usepackage{mathptmx} 
\usepackage{times} 
\usepackage{amsmath} 
\usepackage{amssymb}  
\usepackage{graphicx}
\usepackage{algorithm2e}
\usepackage[export]{adjustbox}
\usepackage{cite}
\usepackage{subfigure}

\title{\LARGE \bf
Local Online Motor Babbling: Learning Motor Abundance of A Musculoskeletal Robot Arm*
}

\author{Zinan Liu$^{1}$, Arne Hitzmann$^{2}$, Shuhei Ikemoto$^{2}$, Svenja Stark$^{1}$, Jan Peters$^{1}$, Koh Hosoda$^{2}$
\thanks{*This project has received funding from the European Union's Horizon 2020 research and innovation programme under grant agreement No 713010 and No 640554}
\thanks{$^{1}$Liu Z. is with Department of Computer Science, TU Darmstadt, Hochschulstr. 10
D-64289 Darmstadt, Germany
        {\tt\small zinan.liu@stud.tu-darmstadt.de}}%
\thanks{$^{2}$Hitzmann A. is with School of Engineering Science, Osaka University, 1-3 Machikaneyama, Toyonaka, Osaka 560-8531 Japan
        {\tt\small arne.hitzmann@arl.sys.es.osaka-u.ac.jp}}%
}

\begin{document}

\maketitle
\thispagestyle{empty}
\pagestyle{empty}

\begin{abstract}
Motor babbling and goal babbling has been used for sensorimotor learning of highly redundant systems in soft robotics. Recent works in goal babbling has demonstrated successful learning of inverse kinematics (IK) on such systems, and suggests that babbling in the goal space better resolves motor redundancy by learning as few sensorimotor mapping as possible. However, for musculoskeletal robot systems, motor redundancy can be of useful information to explain muscle activation patterns, thus the term motor abundance. In this work, we introduce some simple heuristics to empirically define the unknown goal space, and learn the inverse kinematics of a 10 DoF musculoskeletal robot arm using directed goal babbling. We then further propose local online motor babbling using Covariance Matrix Adaptation Evolution Strategy (CMA-ES), which bootstraps on the collected samples in goal babbling for initialization, such that motor abundance can be queried for any static goal within the defined goal space. The result shows that our motor babbling approach can efficiently explore motor abundance, and gives useful insights in terms of muscle stiffness and synergy.
\end{abstract}

\section{INTRODUCTION}
The human body is an over-actuated system, not only does it have a higher dimension in motor space than the degree of freedoms in the action space, i.e., more number of muscles than joints, it also has more degree of freedoms (DoFs) than necessary to achieve a certain motor task. How the effector redundant system adaptively coordinates movements remains a challenging problem. In the field of robot learning, when assuming rigid body links with pure rotation and translation\cite{book}, model learning is commonly used to learn the forward or inverse models of kinematics and dynamics for accurate yet agile control\cite{model}. However for bio-mechanical and soft robots such as the elephant trunks\cite{trunk}, or musculuskeletal systems\cite{anthropomorphic}\cite{biomechanical}, where models based on rigid body links are no longer available, learning becomes difficult due to the highly redundant and non-stationary nature of such systems. 

\begin{figure}
    \centering
    \includegraphics[width=0.75\columnwidth]{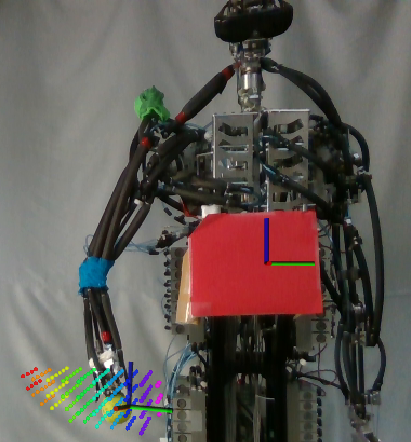}
    \caption{10 DoF musculoskeletal robot arm actuated by 24 pneumatic artificial muscles (PAMs), with an empirically defined goal space in reference to the red marker, visualized in \texttt{rviz}.}
    \label{fig:arm}
    \vspace{-3mm}
\end{figure}

This paper investigates the reaching skills and motor variability of the reached points on a musculuskeletal robot arm\cite{arne}, an over-actuated system of 24 Pneumatic Artificial Muscles (PAMs) actuating 10 DoFs, as shown in Fig. \ref{fig:arm}. Traditionally, this problem could be addressed by learning the forward kinematics using motor babbling, and explore the motor-sensory mapping from scratch\cite{motorbabbling1, motorbabbling2, IK_schaal} until eventually the robot can predict the effects of its own actions. However autonomous exploration without prior knowledge in motor babbling doesn't scale well to high dimensional sensorimotor space, due to the rather inefficient sampling of random motor commands in over-actuated systems. An alternative in \cite{intrinsic} suggests that learning inverse kinematics by goal babbling with active exploration, avoids the curse of dimensionality simply because the goal space is of much smaller dimension than the redundant motor space. Nonetheless, \cite{intrinsic} assumes that the sensorimotor space can be entirely explored, which is not feasible in practice for high dimensional motor systems\cite{trunk}. Another alternative is then to specify the goal space a priori as a grid, and sampling the goal grid points to guide exploration\cite{bootstrapping}, such that sensorimotor mapping can be sufficiently generalized and bootstrapped for efficient online learning. It has also been quantitatively evaluated for an average of sub-centimeter reaching accuracy on an elephant trunk robot \cite{trunk} with reasonable experiment time. We therefore implement and further extend on directed goal babbling in \cite{trunk}. Since the goal space of the robot arm is unknown and non-convex\cite{arne}, we empirically estimate the goal space with randomly generated postures, forcing the convex hull such that directed goal babbling can be applied, and subsequently remove the outlier goals in the goal space after learning.

Given the above works aiming to reduce motor redundancy for learning\cite{motorbabbling1, motorbabbling2, IK_schaal, intrinsic, bootstrapping, trunk}, it can be argued that motor redundancy in human musculoskeletal systems is actually the key stone to natural movements with flexibility and adaptability, hence should be termed motor abundance rather than redundancy \cite{abundance}\cite{bliss}. In robotics motor learning, \cite{variability} also suggests that joint redundancy facilitates motor learning, whereas task space variability does not. Thus we build on directed goal babbling \cite{trunk}, and and propose local online motor babbling, in order to explore motor abundance while fixing space variability on the musculoskeletal robot arm. Local online motor babbling uses CMA-ES initialized by local samples generated from directed goal babbling, such that explorations in the motor space is effectively constrained locally to any queried goal within the goal space, and efficiently generated by adapting the covariance. 


This paper is organized as follows: in Section \ref{goal babbling} and \ref{cmaes} directed online goal babbling and CMA-ES are reviewed. Section \ref{IK} introduces the simple heuristics to define the goal space, implements directed goal babbling on the musculoskeletal robot arm, and evaluates the learning results. Section \ref{motor babbling} proposes, implements, and evaluates local online motor babbling using CMA-ES to query motor abundance, while providing some insights in muscle stiffness and muscle synergy of the musculoskeletal robot system. Section \ref{conclusion} concludes the paper and discusses possible future research directions.

\section{DIRECTED GOAL BALLING}\label{goal babbling}
Given the specified convex goal space $\mathbf{X^*}\in\mathbb{R}^{n}$ encapsulating $K$ goal points, and denoting all the reachable set of commands in the motor space as $\mathbf{Q}\in\mathbb{R}^{m}$, the aim is to learn the inverse kinematics model $\mathbf{X^*}\rightarrow\mathbf{Q}$, that generalizes all points in the goal space to a subset of solutions in the motor space. Starting from the known home position $x^{home}_0$, and home posture $q^{home}_0$, i.e., the inverse mapping $g(x^{home}_0) = q^{home}_0$, the goal-directed exploration is generated by 
\begin{equation}\label{base}
q_t^* = g(x^*_t, \theta_t)+E_t(x^*_t),
\end{equation}
where $g(x^*_t, \theta_{t})$ is the inverse mapping given learning parameter $\theta_t$, and $E_t(x^*_t)$ adds perturbation noise to discover new positions or more efficient motor commands in reaching goals. At every time step, the motor system forwards the perturbed inverse estimate, $x_t, q_t = fwd(q_t^*)$, and the actual $(x_t, q_t)$ samples are used for regression, where prototype vectors and local linear mapping\cite{llm} is used as the regression model, and to monitor the progress of exploration in the defined goal space.

The major part of directed goal babbling is to direct the babbling of the end effector at specified goals and target positions. Each trial of goal babbling is directed at one goal randomly chosen from $\mathbf{X^*}$, and continuous piecewise linear targets are interpolated along the path
\begin{equation}\label{movement}
x^*_{t+1} = x_t^* + \frac{\delta_x}{||X_g^* - x_t^*||}\cdot(x_g^* - x_t^*),
\end{equation}
where $x_{t}^*, X_{g}^*$ are the target position and final goal of the trial, and $\delta_{x}$ being the step size. Target positions are generated until $x_{t}^*$ is closer than $\delta_{x}$ to $X_g^*$, then a new goal $X_{g+1}^*$ is chosen. The purpose of directed goal babbling is to generate smooth movement around the end effector position, such that the locally learned prototype vectors can bootstrap and extend the exploration of the goal space, and allow the integration of the following weighting scheme
\begin{align}
w_t^{dir} &= \frac{1}{2}(1+\arccos(x_t^*-x_{t-1}^*, x_t - x_{t-1}) \\
w_t^{eff} &= ||x_t - x_{t-1}||\cdot||q_t-q_{t-1}||^{-1}  \\
w_t &= w_t^{dir} \cdot w_t^{eff},\label{weight}
\end{align}
$w_t^{dir}$ and $w_t^{eff}$ measure direction and kinematic efficiency of the movement, such that inconsistency of a folded manifold, and redundant joint positions can be optimized\cite{bootstrapping}. The multiplicative weighting factor $w_t$ is then integrated to the gradient descent that fits the currently generated samples by reducing the weighted square error (see Appendix).

To prevent drifting to irrelevant regions and facilitate bootstrapping on the local prototype centers, the system returns to $(x^{home}, q^{home})$ with probability $p^{home}$ instead of following another goal directed movement. Returning to home posture stablizes the exploration in the known area of the sensorimotor space [12], [18], similar to infants returning their arms to a comfortable resting posture between practices:
\begin{equation}
    q^*_{t+1} = q^*_t + \frac{\delta_q}{||q_{home}-q_t^*||\cdot(q^{home} - q_t^*)},
\end{equation}
 the system moves from the last posture $q_t^*$ to the home posture $q^{home}$ in the same way as in (\ref{movement}) by linearly interpolating the via-points along the path, until $||q_{home}-q_t^*|| < \delta_q$

The exploratory noise, or motor perturbation in \ref{base}, is crucial for discovering new postures that would otherwise not be found by the inverse estimate [12], [29]. By exploring the local surrounding of the inverse estimate with i.i.d normal distribution in each motor dimension, and varying these distribution parameters with a normalized Gaussian random walk, the noise is modeled as:
\begin{equation} \label{noise}
    E_{t}(x^*_t) = A_t\cdot x_t^* + b_t, \quad A_t \in \mathbb{R}^{m\times n},\quad  b_t \in \mathbb{R}^{m},
\end{equation}
where all entries $e_t^i$ in the matrix $A_t$ is initialized and varied as follows:
\begin{align*}
    e^i_{0} &\sim \mathcal{N}(0, \sigma^2), \quad \delta_{t+1}^i \sim \mathcal{N}(0, \sigma^2_{\Delta})\\
    e^i_{t+1} &= \sqrt[]{\frac{\sigma^2}{\sigma^2+\sigma_{\Delta}^2}} \cdot (e_t^i + \delta_{t+1}^i)\sim\mathcal{N}(0, \sigma^2).
\end{align*}

After learning, the average reaching accuracy is evaluated by querying the inverse model for every goal within the defined goal space $\mathbf{X^*}$, and a simple feedback controller to adapt execution failures. Execution failure occurs when the inverse estimate is not possible to execute, i.e., $q^* \notin \mathbf{Q}$, due to interference, non-stationary bionic robot design and $\mathbf{Q}$ constantly changing overtime. Given the queried goal $x^*$ and the predicted posture $q^*=g(x^*)$, where $q^* \notin \mathbf{Q}$, the feedback controller would slightly shift the queried goal from $x^*$ to $\hat{x}_t^*$, then forwarding the inverse estimate $x_{t}=fwd(g(\hat{x}^*_t))$. Target shifting follows the current observed error $err_t = x^* - x_t$, and integrated over time:
\begin{equation}
    \hat{x}^*_0=x^*, \quad \hat{x_{t+1}^*}=\hat{x_{t}^*}+\alpha\cdot err_{t}. \label{fb}
\end{equation}

\section{CMA-ES}\label{cmaes}
CMA-ES is a method of black box optimization that minimizes the objective function $f: \mathbf{Q}\in \mathbb{R}^{m} \rightarrow \mathbb{R}$, $q\rightarrow f(q)$, where $f$ is assumed to be a high dimensional, non-convex, non-separable, and ill-conditioned mapping of the multi-variate state space. The idea of CMA-ES is introducing a multi-variate normal distribution to sample a population, evaluating the population $f(\mathbf{q})$ to select the good candidates, and updating the search distribution parameters by adapting the covariance and shifting the mean of the distribution according to the candidates.

Given a start point $q^0$ and initializing the covariance to identity matrix $\mathbf{C^0} = \mathbf{I}$, the search points in one population iteration is sampled as follows:
\begin{equation}
    q_i^t \sim m^t + \sigma^t y_i^t \quad i = 1, \cdots, \lambda \quad q_i, m\in \mathbb{R}^{n}, \sigma \in \mathbb{R}_{+}, \mathbf{C}\in\mathbb{R}^{n\times n}\label{sample}
\end{equation}
where $y_i^t=\mathcal{N}_i(\mathbf{0}, \mathbf{C^t})$, $m$ being the mean vector, $\sigma$ being the step-size, and $\lambda$ is the population size. For notation simplicity, the iteration index $t$ is henceforth omitted.

The mean vector $m$ is updated by using the non-elitistic selection \cite{tutorial}. Let $q_{i:\lambda}$ denote the $i$th best solution in the population of $\lambda$, the best $\mu$ points from the sampled population are then selected, such that $f(q_{1:\lambda})\leq\cdots\leq f(q_{\mu:\lambda})$, and weighted intermediate recombination is applied:
\begin{equation}
m \leftarrow m + \Sigma_{i=1}^{\mu}w_{i} y_{i:\lambda} =: m + y_w, \label{update mean}
\end{equation}
$$
\text{where}\;
w_1 \geq\cdots\geq w_{\mu}>0, \; \Sigma_{i=1}^{\mu} w_{i}=1, \; \frac{1}{\Sigma_{i=1}^{u}w_{i}^2}=:\mu_{w}\approx \frac{\lambda}{4}.
$$
The step size $\sigma$ is updated using cumulative step-size adaptation (CSA). The intuition is when the evolution path, i.e., the sum of successive steps, is short, single steps tend to be uncorrelated and cancel each other out, thus the step-size should be decreased. On the contrary, when evolution path is long, single steps points to similar directions and tend to be correlated, therefore increasing the step size. Initializing the evolution path vector $p_{\sigma} = \mathbf{0}$, and setting the constants $c_{\sigma}\approx 4/n, d_{\sigma}\approx 1$, the step size is updated as: 
\begin{align}
p_{\sigma} & \leftarrow (1-c_{\sigma}) p_{\sigma}+ \sqrt{1 - (1-c_{\sigma})^2}\;\sqrt{\mu_w}\;y_w\label{update_path_s}\\ 
\sigma & \leftarrow \sigma \times \exp{\left(\frac{c_{\sigma}}{d_{\sigma}}\left(\frac{||p_{\sigma}||}{E||\mathcal{N}(\mathbf{(0, I)})||}-1\right)\right)}\label{update sigma}
\end{align}
The essential part of the evolution strategy is the covariance matrix adaptation. It is suggested that the line distribution adapted using rank-one update will increase the likelihood of generating successful steps $\mathbf{y_w}$, because the adaptation follows a natural gradient approximation of the expected fitness of the population $f(\mathbf{q})$.
\begin{align}
    p_c &\leftarrow (1-c_c) p_c + \sqrt{1-(1-c_c)^2}\;\sqrt{\mu_w}\;
    y_w\label{update_path_c}\\ 
    \mathbf{C} &\leftarrow (1-c_{cov})\mathbf{C} + c_{cov} p_c p_c^{T}\label{update C}
\end{align}

\section{INVERSE KINEMATICS LEARNING}\label{IK}
We use a 10 DoF musculoskeletal robot arm from \cite{arne} for the experiments, where the robot is controlled via ROS messages. The arm is driven by 24 pneumatic muscles, each with pressure actuation range of $[0, 0.4]$MPa.
 \begin{figure}
    \centering
    \includegraphics[width=1\columnwidth, inner]{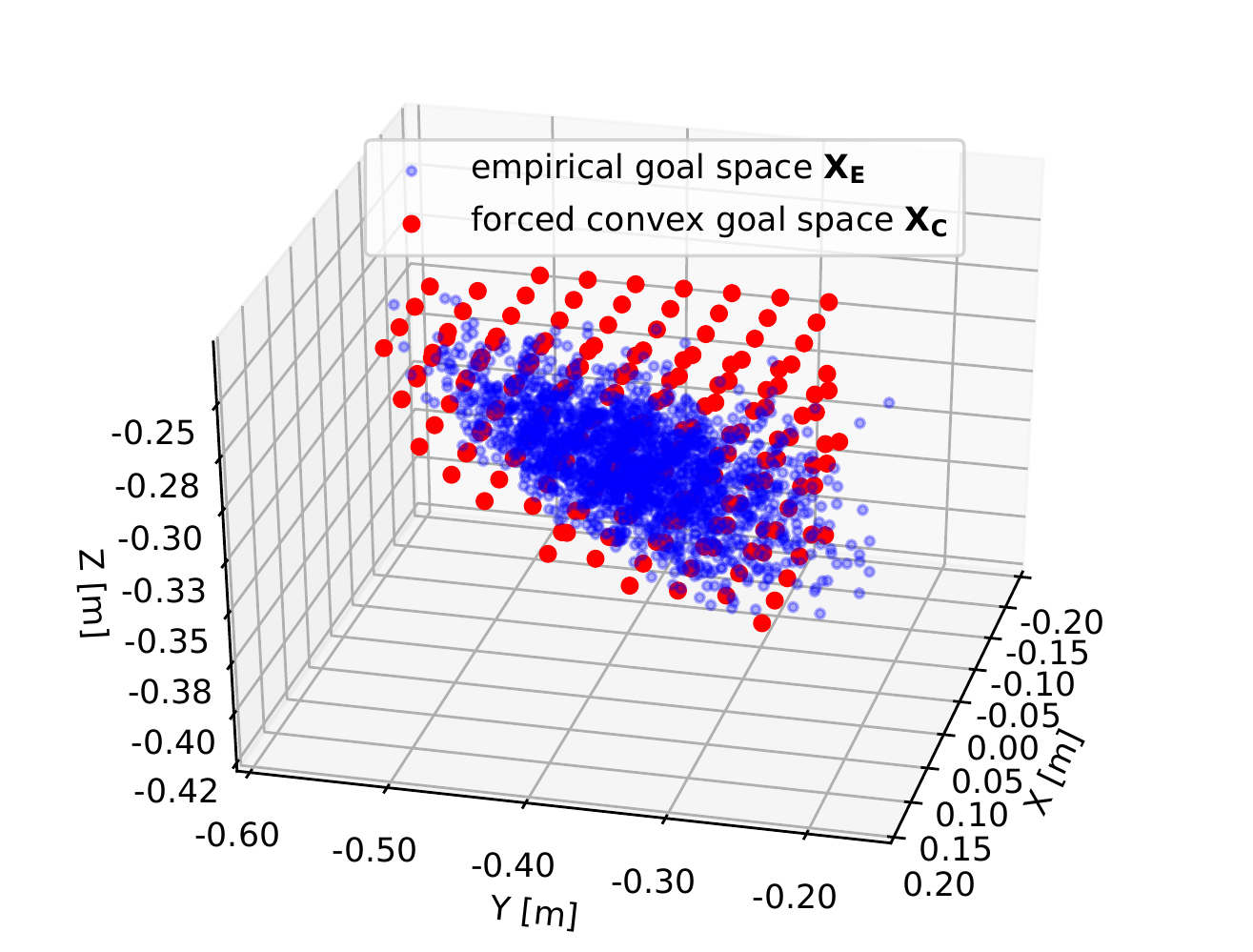}
    \caption{Empirical goal space $\mathbf{X_E}$(in blue) sampled from 2000 random postures, and the convex goal space $\mathbf{X_C}$(in red), which is used for learning as shown in Fig. \ref{fig:arm}}
    \label{fig:goalspace}
    \vspace{-3mm}
\end{figure}
 As shown in Fig.~\ref{fig:arm}, the hand of the robot is replaced with a tennis ball as the color marker, and tracking of the end effector is performed in reference to the center of the red marker as the origin, using Intel RealSense ZR300. However the tracking introduces an error up to 1cm in depth, i.e., x-axis, and sub-millimeter error in y and z axis. The colored point cloud overlayed in ROS rviz is the specified convex goal space as in Fig. \ref{fig:arm}. The control accuracy of the robot is tested according to \cite{trunk}. By repeating $P=20$ random postures for $R=20$ times each, the average Euclidean norm error is computed to be $D=1.2$ cm as follows:
\begin{align*}
    \Bar{x}_p & = \frac{1}{R}\sum_{r}x_{p}^r \\
    D & = \frac{1}{P}\sum_{p}\frac{1}{R}\sum_{r}||x_p^r-\Bar{x}_p||
\end{align*}

\subsection{Define the Goal Space}
The complete task space of the upper limb robot is unknown and non-convex, however directed goal babbling would require the specified goal space to be convex to efficiently bootstrap and allow the integration of the weighting scheme in (\ref{weight}). Thus we first empirically estimate the goal space by randomly generating $2000$ random postures for each muscle within $[0, 0.4]$ MPa, and take the encapsulated convex hull as the empirical goal space $X_E$. In order to approximate the uniform samples in $X_E$ for efficient online learning and evaluations,  a cube grid $\mathbf{C}$ with $3cm$ spacing encapsulating $\mathbf{X}_E$ is defined, where  $\mathbf{X}_{E} \subset \mathbf{C}$. The sampled convex hull goal grid $\mathbf{X_C}$ in Fig. \ref{fig:arm} is then made from the intersection of all points in the empirical goal space and the cube grid, i.e., $\mathbf{X_C} = \mathbf{X}_E \cap \mathbf{C}$. However, as shown in Fig. \ref{fig:goalspace}, $\mathbf{X}_E$ is a slanted non-convex irregular ellipsoid, forcing a convex hull in the $2000$ random posture samples would introduce non-reachable regions in the goal space. This is addressed later with the similar set operation to remove the outlier goals using the learned prototype vector space.
 
 \begin{figure}
    \centering
    \includegraphics[width=0.8\columnwidth]{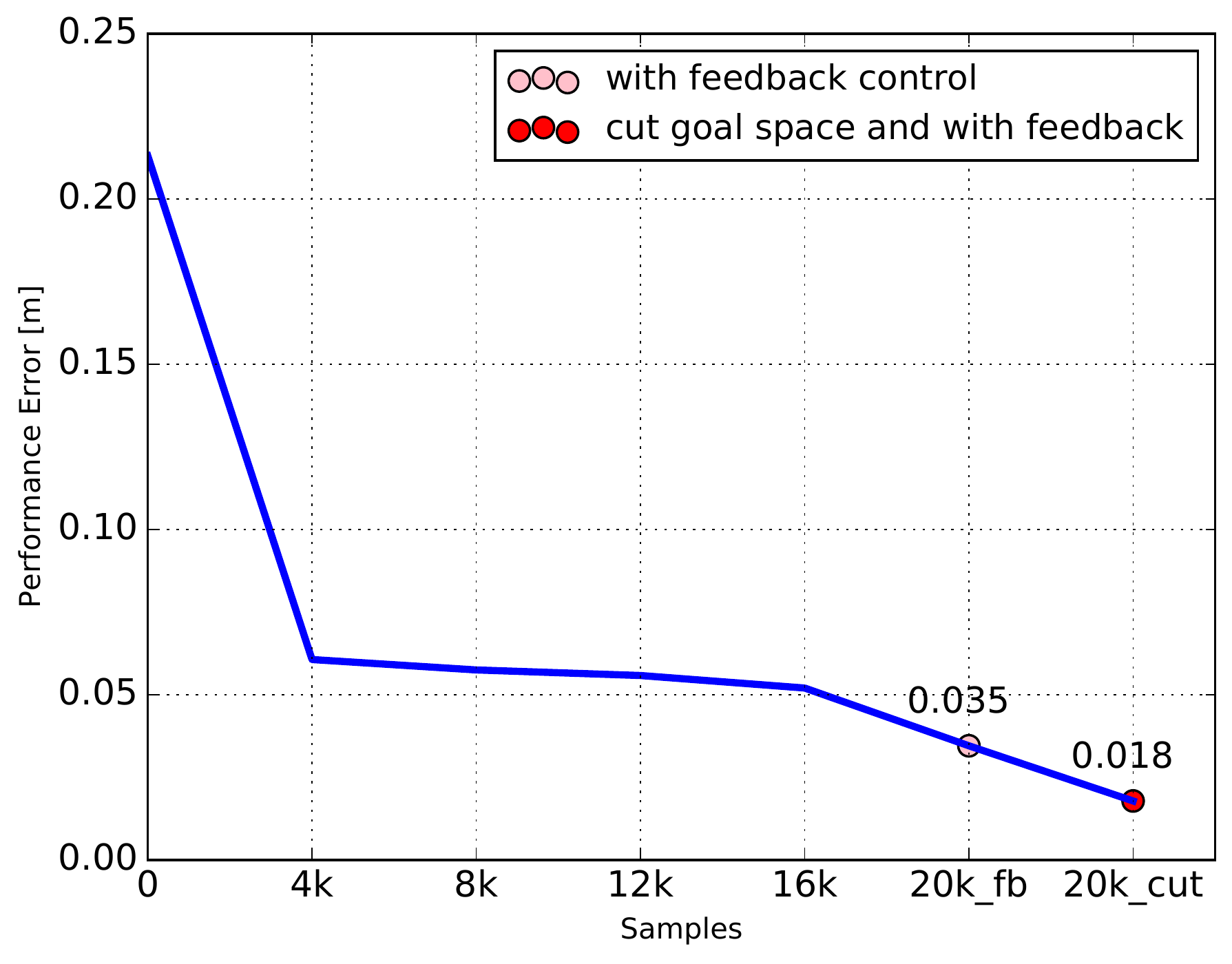}
    \caption{Decreasing performance error up to 20000 samples, i.e., the average Euclidean norm to the goals evaluated throughout the convex goal space $\mathbf{X_C}$, the feedback controller is applied at 20000 samples, resulting an average error of 3.5 cm. However there are still outlier goals remaining from forcing the convex hull, thus we take the explored prototype sphere space $\mathbf{S}$ and intersect with $\mathbf{X_C}$, i.e., $\mathbf{X_S}=\mathbf{S} \cap \mathbf{X_C}$ to remove the outliers. Evaluating on the cut goal space $\mathbf{X_S}$ reduces the error to 1.8 cm}
    \label{fig:error curve}
\end{figure}

 \begin{figure}
    \centering
    \includegraphics[width=0.8\columnwidth]{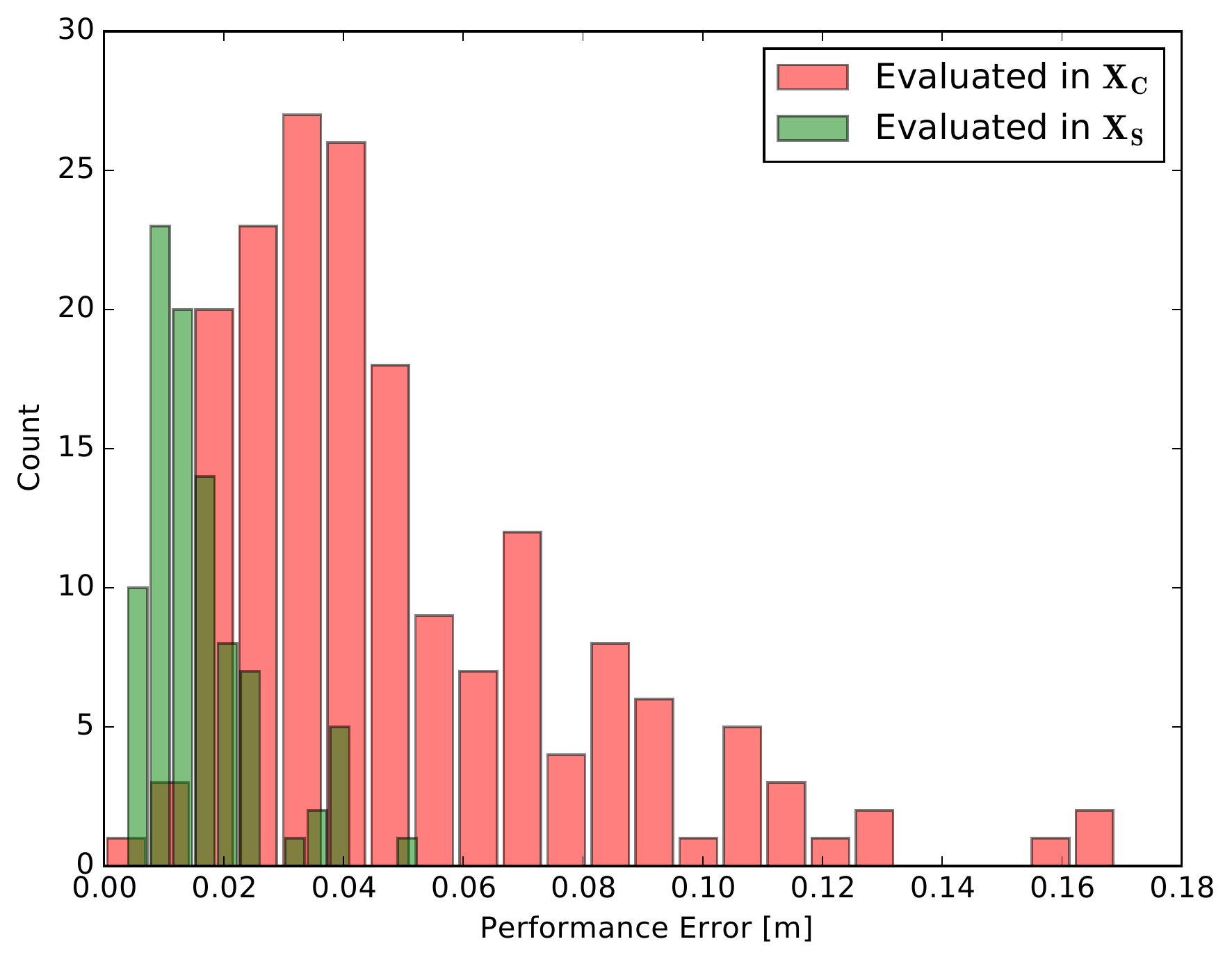}
    \caption{Performance error distribution of the convex goal space $\mathbf{X_C}$ used for IK learning in Fig. \ref{fig:goalspace}, and that of the cut goal space $\mathbf{X_S}$ in Fig. \ref{fig:final goalspace}}
    \label{fig:hist}
\end{figure}

\subsection{Experiment and Results}
\begin{figure}
    \centering
    \includegraphics[width=1\columnwidth, inner]{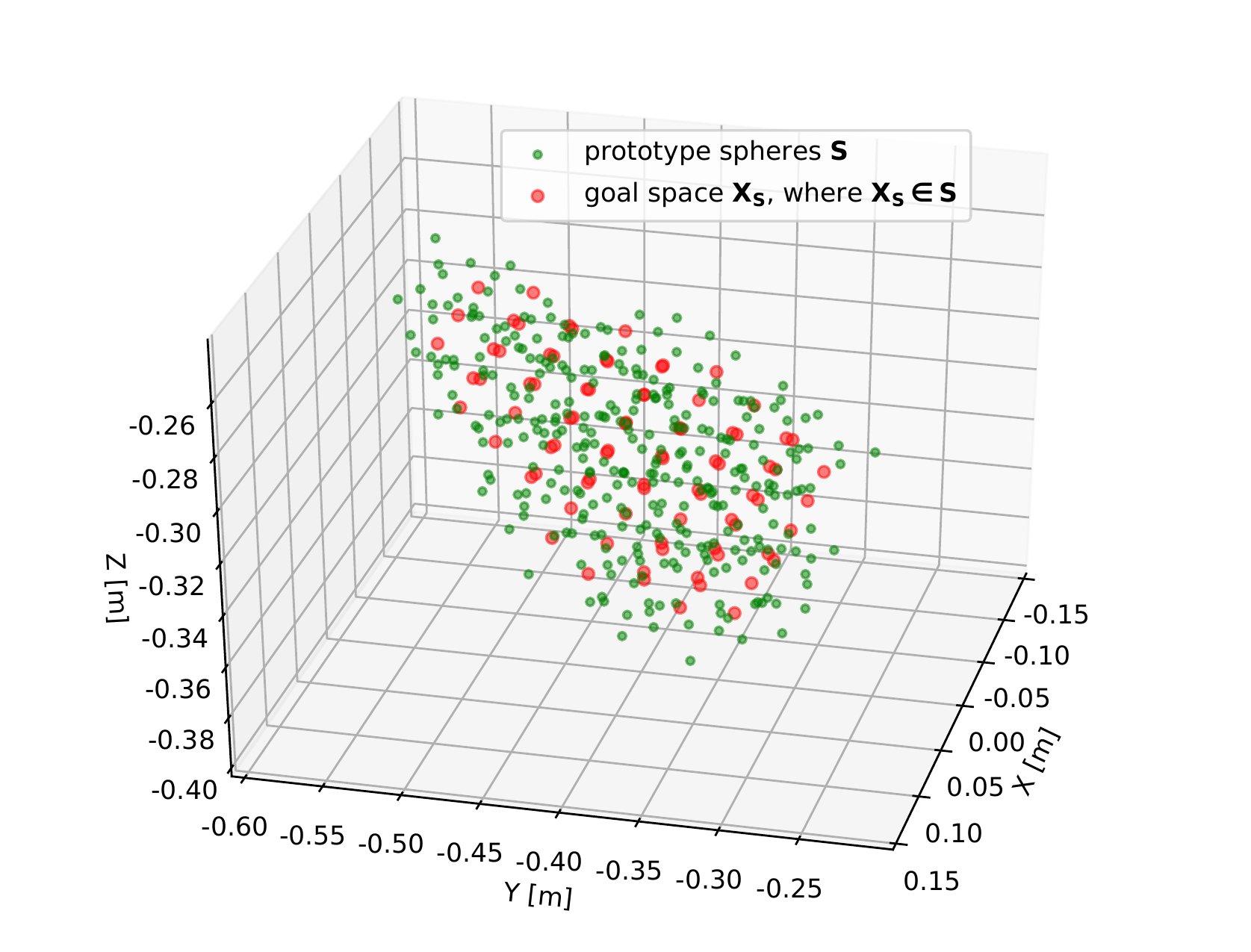}
    \vspace{-5mm}
    \caption{Prototype spheres $\mathbf{S}$ (green) encapsulating the final goal space $\mathbf{X_S}$ (red), which is used for goal babbling, after outlier goals has been removed.}
    \label{fig:final goalspace}
    \vspace{-5mm}
\end{figure}

The experiment is conducted with $T=20000$ samples, with target step length $\delta_{x}=0.02$, which corresponds to the target velocity of $2$ cm/s, allowing the robot to generate smooth local movements. The sampling rate is set to $5 Hz$, generating 5 targets and directed micro movements for learning. After every 4000 samples, performance evaluation is carried out online. The learning experiment including online evaluations amount to less than 2 hours real time. As illustrated in Fig. \ref{fig:error curve}, the learning bootstraps quite fast in the first 4000 samples, followed by a slow convergence until 16000 samples. At $T=20000$, the feedback controller is applied, the performance error drops to 3.4cm. However in $\mathbf{X_C}$ there are still many outlier goals, which are the non-reachable regions introduced by forcing the convex hull. A similar set intersection operation is applied with the learned prototype spheres $\mathbf{S}$ and the goal space $\mathbf{X_C}$, where $\mathbf{S}$ is taken as the encapsulated space of the prototype spheres, and the final goal space is $\mathbf{X_S}=\mathbf{S} \cap \mathbf{X_C}$, as shown in Fig. \ref{fig:final goalspace}, where the number of goals has been reduced from 179 in $X_C$ to 94 in $X_S$. We then evaluate again these 94 goals with the feedback controller, the performance error reduces further to an average of 1.8 cm in \ref{fig:error curve}. However due to the forced convex hull $X_C$, local inverse models cannot efficiently regress at the edge of the task space, the error distribution still shows a few errors larger than 3 cm, which can be further reduced later by motor babbling using CMA-ES.

\section{LEARNING MOTOR REDUNDANCY}\label{motor babbling}
CMA-ES explores by expanding the search distribution of the parameters, shifting the mean and expanding covariance, until optimum solution is found within that distribution, followed by shrinking the covariance and shifting the mean to the global optimum. By intentionally setting the initial mean vector slightly away from the opitimum, i.e., the posture that leads to closest end effector position to the goal, CMA-ES would naturally expand the covariance while keep the search within the vicinity of the queried goal, as the objective function is set to minimize the goal reaching error. Essentially CMA-ES is used to effectively generate motor babbling data, which can be achieved by initializing the mean vector with neighboring goal postures $g(\hat{x})$ of the queried goal $x^*$, and the step-size with the empirical variance estimate of the local samples around $x^*$ gathered from the goal babbling process.

\subsection{Local Online Motor Babbling}
When learning inverse kinematics using online goal babbling, since there are multiple postures $\mathbf{q}$ reaching $x^*$, it is assumed that we don't need to know all redundancy, and only learn the ones with most direction and kinematic efficiency by integrating the weighting scheme (\ref{weight}) in the optimization. In fact, $\mathbf{Q}$ is not only unknown, and may never be exhaustively explored on a physical system, but also non-stationary due to the nature of musculoskeletal robot design with PAMs. This can be addressed by using the simple feedback controller in (\ref{fb}), where execution failures due to the changing of $\mathbf{Q}$ are adapted when the queried goal $x^*$ is slightly shifted based on the proportion of the euclidean error $x^*-x_t$.

\RestyleAlgo{boxruled}
\begin{algorithm}
\SetKwInOut{Input}{input}\SetKwInOut{Output}{output}\SetKwInOut{Initialize}{initialize}
\Input{$x^*$, $g(x)$, $\mathbf{Q_{\hat{x}}}$}
\Output{$\mathbf{Q_{cma}}$}
\Initialize{$\alpha=0.05,\; T=30,\; N=5,\; \lambda=13,\; r=0.02,\; c=10,\; f^*=0.03,\; \mathbf{Q_{cma}=\{\}}$}
\BlankLine
\emph{select N closest goals $x^*_1, \cdots, x^*_N$ to $x^*$}\;
\For{$n\leftarrow 0$ \KwTo $N$}{
$\hat{x}^*_0 = x^*_n$\;
$\mathbf{Q_{fb}}=\{\}$\;
 \For{$t\leftarrow 0$ \KwTo $T$}{
    $x_t, q_t = \text{forward}(g(\hat{x}^*_t))$\;
    $\hat{x}^*_t = \hat{x}^*_{t-1} + \alpha \cdot (x^* - x_t)$\;
    \If{$||x^* - x_t||<r$}{collect ($x_t$, $q_t$) In $\mathbf{Q_{fb}}$\;}
    }
\emph{select $q_{t}$ for the minimum $||x_{t}-x^*||$ in $\mathbf{Q_{fb}}$}\;
\emph{initialize $m = q_{t},\; \sigma=\text{mean(var}(\mathbf{Q_{\hat{x}}}\cup \mathbf{Q_{fb}})),\; \mathbf{C} = \mathbf{I}$}\;
 \While{$\hat{f}<f^*$}{
    sample posture population $\mathbf{q_s}: q_1\cdots q_{\lambda}$ as in (\ref{sample})\;
    \For{$k\leftarrow1$ \KwTo $\lambda$}{
        $x_t, q_t = \text{forward}(q_k)$\;
        $\hat{f} = f(x_k) = c\cdot ||x^* - x_t||$\;
        \If{$||x^* - x_t||<r$}{
            collect $q_k$ in $\mathbf{Q_{cma}}$\;}
        }
    update $\mathbf{m}$ as in (\ref{update mean})\;
    update $\mathbf{p_{\sigma}}$ and $\sigma$ as in (\ref{update_path_s}), (\ref{update sigma})\;
    update $\mathbf{p_{c}}$ and $\mathbf{C}$ as in (\ref{update_path_c}), (\ref{update C})\;
    }
}\caption{Motor Babbling Using CMA-ES}\label{algo}
\end{algorithm}

As illustrated in Algorithm \ref{algo}, the queried goal $x^*$, the learned inverse model $g(x)$, and the neighboring postures $\mathbf{Q_{\hat{x}}}: q_t \forall x_t \iff ||x_t-x^*||<r$, which is collected from the goal babbling process, are the input to online motor babbling. The aim of the algorithm is to output a new posture configuration set $\mathbf{Q_{cma}}$, from which different muscle stiffness can be generated while keeping the end effector position fixed.  The initialization sets the gain and number of iteration of the feedback controller to $\alpha=0.05$, $T=30$, t number of trials for CMA-ES $N=5$, and the prototype sphere radius is $r=0.02$. We use \texttt{pycma} library\cite{pycma} to implement CMA-ES, where we encode variables $q$ in the objective function implicitly $f(fwd(q))$\cite{tutorial}. The objective function is simply set as the euclidean norm to the goal scaled with a constant, i.e., $c\cdot||x^*-x_t||$, where $c=10$, and the optimum objective function value is set to $f^*=0.03$, meaning that an empirical optimum of $f^*/c=3 mm$ to the goal, which is also the stopping criteria for each CMA-ES trial.

Each trial of CMA-ES starts by iterating the feedback controller and finding the posture $q_t$ that leads closest to the neighboring goal, and $q_t$ is subsequently used to initialize the mean vector $m$. The covariance is initialized to be an identity matrix, which allows isotropic search and avoids bias. In order to initialize the step-size, an empirical variance is estimated from $\mathbf{Q_{\hat{x}}}\cup \mathbf{Q_{fb}}$, and the mean of the variance is taken as initialization. The union of the two sets is to ensure sufficient data for a feasible estimation. Near the home position, which is the centroid of the goal space, many data samples are available as online goal babbling often comes back to $(x_{home}, q_{home})$. However around the edges of the goal space, there are often very few local samples, sometimes less than the action space dimension, i.e., the 24 muscles. By taking in the samples generated by feedback controller, a better initialization of $\sigma$ can be robustly estimated. 

\subsection{Visualizing Muscle Abundance}
In order to visualize muscle abundance, namely in terms of reproducing muscle stiffness and muscle synergy encoded in the evolved covariance matrix, we assume the distribution of parameters to be multi-variate Gaussian and multi-modal, as the motor space is of high dimension, and there can be different muscle group posture configurations while keeping the end effector fixed. Therefore a multi-variate Gaussian Mixture Model \cite{bishop} is fit to the collected data in $\mathbf{Q}$. By assuming a distribution of Gaussian parameters over the data samples $p(\mathbf{Q}|\mathbf{\theta})$, a prior multi-variate Gaussian distribution is introduced 
$$p(\mathbf{\theta}) = \Sigma_{i=1}^{K} w_{i}\mathcal{N}(\mathbf{\mu_i}, \boldsymbol{\Sigma_{i}}),$$
$ w_{i}$ are the weights for each Gaussian mixture component, and the posterior distribution is estimated by using Bayes rule \cite{bishop}, such that the posterior distribution would preserve the form Gaussian mixture model, i.e., 
$$p(\mathbf{\theta}|\mathbf{Q}) = \Sigma_{i=1}^{K} \Tilde{w_{i}}\mathcal{N}(\mathbf{\Tilde{\mu_i}}, \boldsymbol{\Tilde{\Sigma_{i}}}),$$
where the parameters $(\mathbf{\Tilde{\mu_i}}, \boldsymbol{\Tilde{\Sigma_{i}}})$ and weights $ \Tilde{w_{i}}$ are updated using Expectation Maximization (EM) to maximize the likelihood \cite{bishop}. The number of mixture models $P$ is estimated using Bayesian Information Criterion (BIC) \cite{bishop} for $P\in [1, 10]$, where the lowest BIC of $P$ is taken. Finally, we sample from the mixture model with updated parameters and weights $q^* \sim \Sigma_{i=1}^{K} \Tilde{w_{i}}\mathcal{N}(\mathbf{\Tilde{\mu_i}}, \boldsymbol{\Tilde{\Sigma_{i}}})$ and forward $q^*$ on the robot.

\subsection{Experiment and Results}
 \begin{figure}
    \centering
    \includegraphics[width=1\columnwidth]{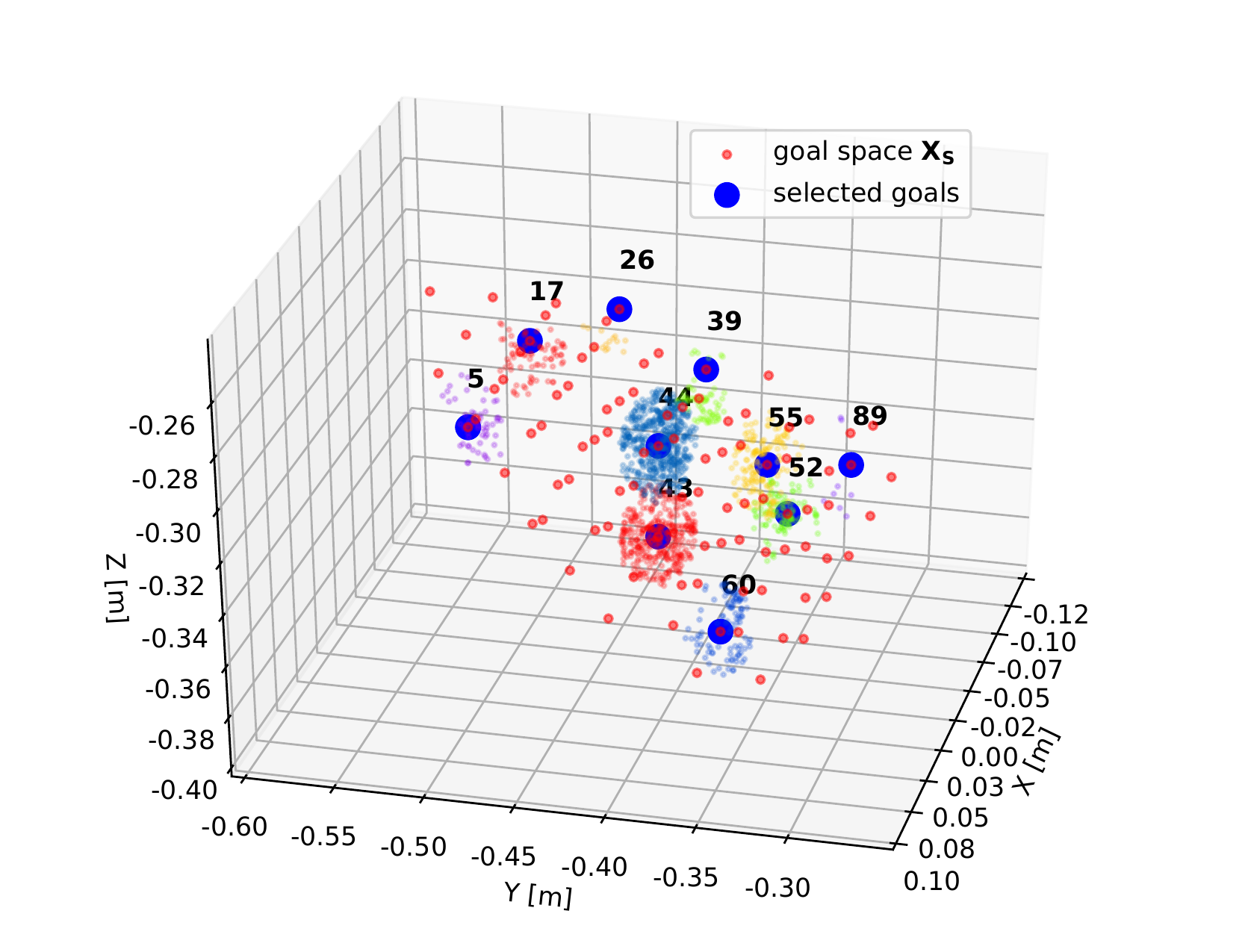}
    \caption{10 selected goals for motor babbling in the final goal space $\mathbf{X_S}$, the color point clouds are the local samples within 2cm radius of the queried goals, which is used to initialize the step-size $\sigma$ for the CMA-ES trials}
    \label{fig:select goal}
\end{figure}
We evenly selected 10 goals in the final goal space $\mathbf{X_S}$ to perform online motor babbling. The selected goals and their local samples within 2cm radius are shown in Figure \ref{fig:select goal}. The goals are selected to show case the generality of querying any goal within the goal space for motor babbling. Around the edges, goal 26, 5, 89, 52 are chosen, and near the centroid home position, goal 44 and 39 are selected. The rest goals 17, 43, 55, and 60 are to populate the rest of the goal space. It can be expected and observed that more samples were generated near the home posture, since in online goal babbling the arm returns to $(x_{home}, q_{home})$ with probability $p_{home}$, whereas goals around the edges have only a few samples, such as goal 26 and 89.

\begin{figure}
    \centering
    \includegraphics[width=1\columnwidth]{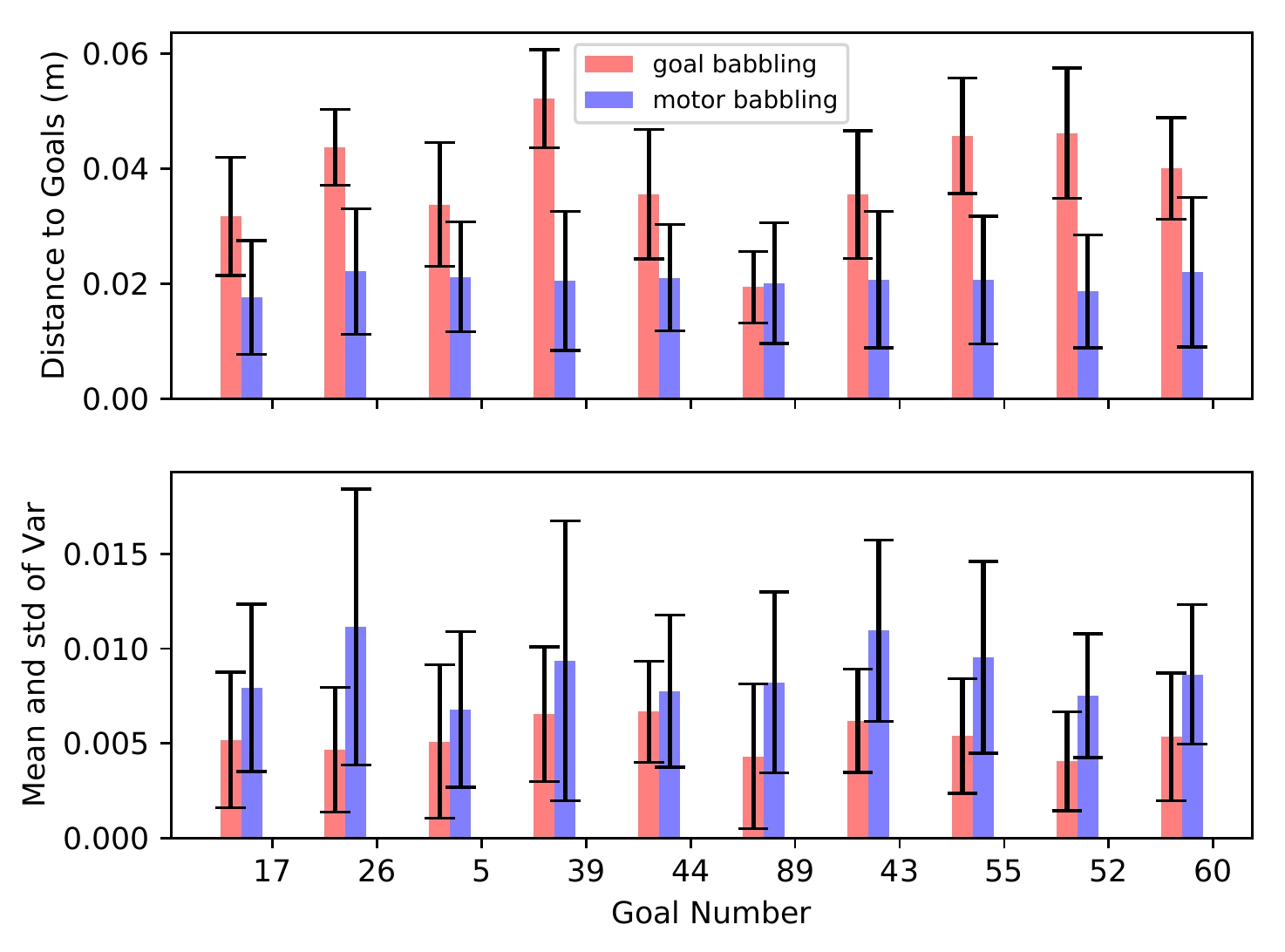}
    \caption{Comparing the reaching error and muscle variability of directed goal babbling and local motor babbling using CMA-ES, where CMA-ES not only increased the means and standard deviations of all 24 muscle variances for the 10 queried goals, but the reaching error has also been reduced}
    \label{fig:final plot}
\end{figure}

For each selected goal, $N=5$ trials of CMA-ES is performed as in Algorithm \ref{algo}, where each trial takes on average 5 minutes experiment time on the robot. Muscle stiffness is then reproduced by first fitting the collected neighboring samples $\mathbf{Q_{\hat{x}}}$ to the Gaussian mixture model, which serves as a baseline learned during goal babbling, followed by another experiment fitting $\mathbf{Q_{cma}}$ to the mixture model and the subsequent sampling. 200 samples from the mixture model is evaluated on the robot, the mean and standard deviation of the reaching error, and of pressure variance are plotted. As illustrated in Fig. \ref{fig:final plot}, CMA-ES outperforms the baseline in terms of both larger muscle pressure variance and smaller goal reaching error, where the average lies close to the 2cm prototype sphere radius. Since online goal babbling favors kinematic and direction efficiency by reducing motor redundancy, the sampled muscle pressure generally varies trivially compared to the ones generated from the CMA-ES GMM model, which expands the variance in search of global optimum while keeping the goal reaching accuracy. Due to non-stationary changes of the possible posture configurations $\mathbf{Q}$, the local neighboring samples $\mathbf{Q_{\hat{x}}}$ no longer lead to a close position to the goal, however $\mathbf{Q_{\hat{x}}}$ of the neighboring goals can be used for initializing the step-size $\sigma$, and initializing the mean vector $m$ from $\mathbf{Q_{fb}}$, to adapt to non-stationary changes and maintain the goal reaching accuracy while performing motor babbling.

\begin{figure}
    \centering
    \includegraphics[width=1\columnwidth]{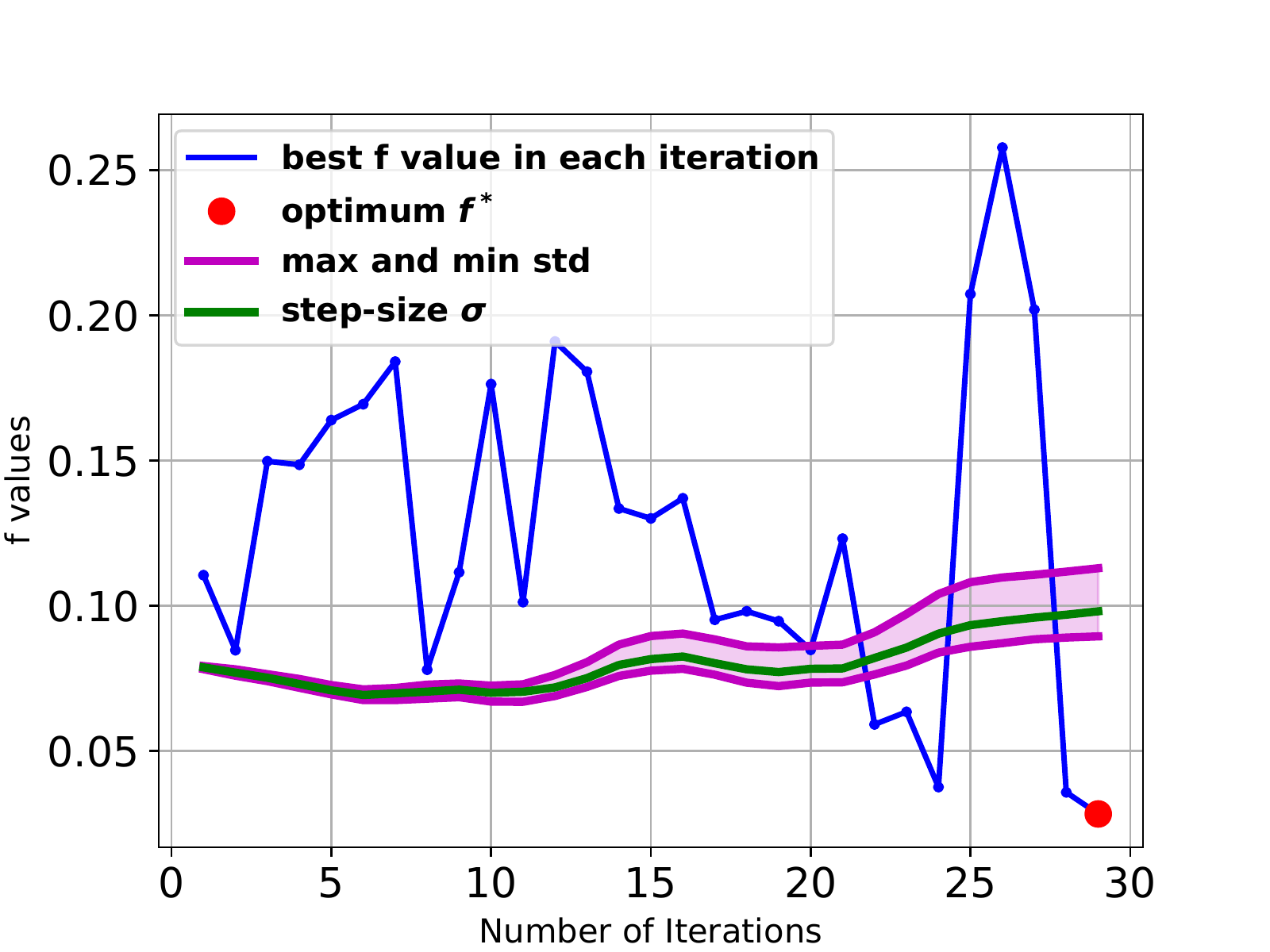}
    \caption{One evolution trial for goal 44, the search of the step-size increases until the defined optimum objective function value is found}
    \label{fig:divers}
\end{figure}

\begin{figure}
    \centering
    \includegraphics[width=0.9\columnwidth]{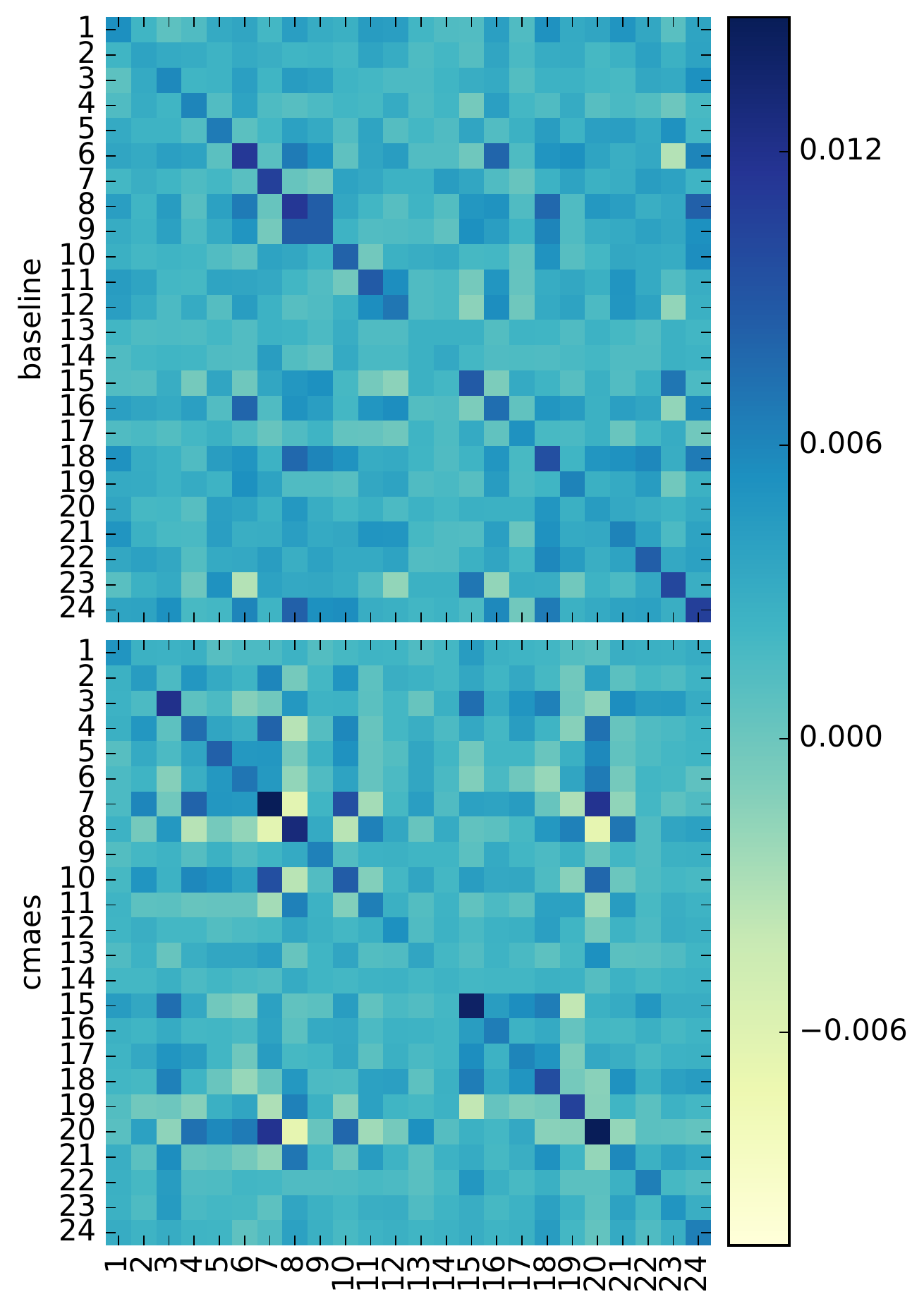}
    \caption{Comparing baseline and CMA-ES covariances, where the largest change of variance occurs at muscle pair (8,20), changing from 0.003 to -0.01, where the -0.01 covariance corresponds to the standard deviation of 0.1 MPa pressure change, consituting 25\% of the PAM actuation range}
    \label{fig:cov}
\end{figure}

It can be observed that for goal 44, which is closest to the home position, the motor variance doesn't increase much as other queried goals. This is because every time the interpolated directed goal path comes across the centroid home region, goal 44 has a higher chance of collecting more samples $q_t$ of varied motor configurations within the neighborhood. Nevertheless, CMA-ES still explores motor redundancy rather efficiently. As shown in Fig. \ref{fig:divers}, the evolution trial expands the maximum and minimum standard deviation of the search, i.e., such that the optimum $f^*$ is reached. After 5 such evolution trials, the sampled GMM data is used to estimate the covariance, compared with the covariance estimate from the baseline GMM data. As shown in \ref{fig:cov}, CMA-ES preserves the structure while enhancing the variance on the diagonal, while also discovers more correlation within different groups of muscles, which can be prominently observed on the robot in Fig. \ref{fig:pics}. 

\subsection{Interpreting Muscle Abundance}\label{synergy}

\begin{table}[h!]
\centering
\begin{tabular}{||c | c | c ||} 
 \hline
\# & Muscle Name & Function \\ [1ex] 
 \hline\hline
    3 & serratus anterior & pulls scapula forward \\ \hline
    7 & latissimus dorsi & rotates scapula downward \\ \hline
    8 & rear deltoid & abducts, flexes, \\
    10 & front deltoid &  and extends \\
    11 & medial deltoid & the shoulder \\ \hline
    15 & biceps brachii & flexes and supinates the forearm \\ \hline
    18 & brachialis & flexes the elbow \\ \hline
    19 & pronator & pronates the hand \\ \hline
    20 & supinator & supinates the hand \\ \hline
\end{tabular}
\caption{Muscle Names and Functions}
\label{table:muscles}
\vspace{-3mm}
\end{table}

The muscle pressure variability in the covariance encodes muscle abundance, which can be interpreted as muscle stiffness and static muscle synergies. Loosely speaking, muscle synergy is defined as a co-activation pattern of muscles in a certain movement from a single neural command signal\cite{synergy}. It can be argued that muscle synergy is a way of kinetically constraining the redundant motor control of limited DoFs, or as neural strategies to handle the sensorimotor systems\cite{vs}. 

\begin{figure}%
\centering
\subfigure[][]{%
\label{fig:1_front}%
\includegraphics[height=5.5cm, width=3.5cm]{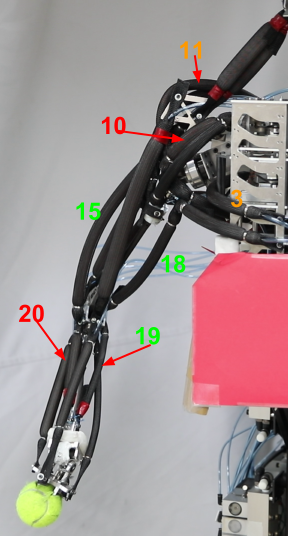}}%
\hspace{8pt}%
\subfigure[][]{%
\label{fig:2_front}%
\includegraphics[height=5.5cm, width=3.5cm]{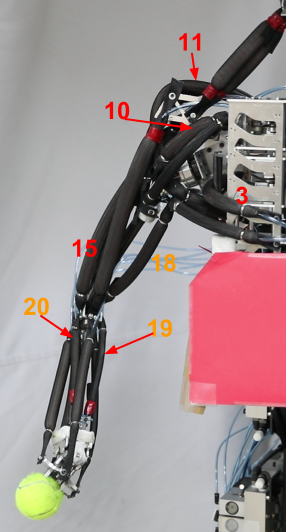}} \\
\subfigure[][]{%
\label{fig:1_back}%
\includegraphics[height=5cm, width=3.5cm]{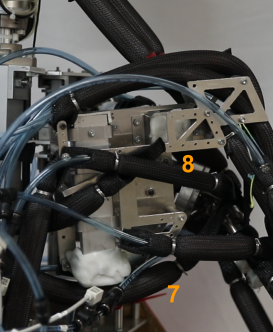}}%
\hspace{8pt}%
\subfigure[][]{%
\label{fig:2_back}%
\includegraphics[height=5cm, width=3.5cm]{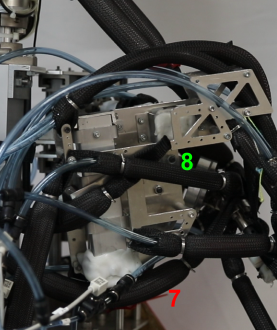}}%
\caption{Static muscle synergies, the labelled muscles are color coded in green (low), orange (medium), and red (high) to indicate the state of pressure actuation. A relaxed arm posture with a lowered shoulder can be observed in (a) and (c), whereas a stiffened arm with pronating hand and a lifted shoulder and can be observed in (b) and (d), while keep the end effector position fixed.}
\label{fig:pics}%
\end{figure}

We claim no sides in the sensorimotor learning of humans, however, by constraining the end effector position of the musculoskeletal robot arm, the static muscle synergies and stiffness can be encoded in the covariance matrix and provide some useful insights. 
In Fig. \ref{fig:cov}, muscles of high variances, namely muscle 3, 7, 8, 10, 11, 15, 18, 19, 20 are of particular interest, where muscle 7 and 8, 20 and 8 are highly negatively correlated. Inspired by the human's upper limb, the PAMs of the robot arm mimics the function of human arm muscles, as illustrated in Table \ref{table:muscles}. By fitting the data $\mathbf{Q_{cma}}$ in the mixture models and subsequently applying sampling, we can observe the co-activation patterns of the muscles. As shown in Fig.\ref{fig:1_front} and \ref{fig:1_back}, the upper limb first reaches goal 44 with a relaxed arm posture and a lowered adducted shoulder, whereas in Fig.\ref{fig:2_front} and \ref{fig:2_back} the end effector position is maintained by stiffening the arm, lifting the extended shoulder, and pronating the hand. The negative correlation of muscle 7 and 8 can be interpreted as the coordination of extension and abduction, as well as the flexion and adduction of the shoulder. Muscle 8 and 20 coordinate shoulder abduction with a supinating hand, and by adducting shoulder while pronating the hand. 

\section{CONCLUSIONS}\label{conclusion}
We have implemented directed goal babbling\cite{bootstrapping} to learn the inverse kinematics of a 10 DoF musculoskeletal robot arm actuated by 24 PAMs. We defined the goal space by empirically sampling 2000 random postures and forcing a convex hull ready for learning, and post-processed the goal space to removed outlier goals. The result shows an average reaching error of 1.8 cm, where the reaching accuracy achievable by the robot is 1.2 cm. The simple heuristics and approximation of the goal space allows us to use directed goal babbling to learn a larger sensory space compared to a well-defined yet small partial task space, and promote more efficient mapping to the motor space compared to active exploration\cite{intrinsic}. Nevertheless, learning with a forced convex goal space where the intrinsic task space is non-convex introduces outlier goals, which the corresponding directed babbling can be misleading. A future research direction of integrating directed goal babbling with active exploration could be of interest, where the goal space grid can be defined large enough to encapsulate the whole task space, and active exploration guided by the k-d tree splitting and progress logging can indicate the learned task space while still keep the bootstrapping flavor of the inverse model.

We further extended directed goal babbling to local online motor babbling using CMA-ES in search of more motor abundance. By initializing the evolution strategy with local samples generated from goal babbling, any point within the goal space can be queried for motor abundance. The idea is to intentionally initializing the mean vector of CMA-ES slightly away from the queried goal. By expanding covariance and setting the stop condition to meeting the set optimum of the objective function value, efficient motor babbling data can be generated locally around the queried goal with a few CMA-ES trials of different initializations from the neighboring goals. We evenly selected 10 goals within the goal space to showcase the generality of local online motor babbling. The results show that our proposed method has significantly increased the average muscle pressure variances, while keeping the end effector more stable and closer to the queried goals, compared with the goal babbling baseline. Even in the home position where motor abundance has already been well-explored, local motor babbling shows a maximum increased standard deviation of 0.1 MPa, constituting 25\% of the muscle pressure actuation range. Our method also adapts to queried goals near the edges of the goal space where samples for initialization are sparse due to the uneasy posture of the robot arm around such goals. 

By fitting Gaussian mixture models to the data collected using local motor babbling, the sampling of the GMMs can reproduce motor abundance in terms of muscle stiffness and muscle synergies encoded in the evolved single-mode covariance matrix. Muscle stiffness can be seen on the inflating and deflating muscles, and muscle synergies can be clearly observed in the covariance where variances and correlations are strong, as well as when GMM sampled postures are applied on the robot correspondingly. The bonus that comes with the encoded covariance and mixture models is that the queried motor abundance can be captured and reproduced by distributions, which enables the formulation of trials for reinforcement learning in future research, such as learning weight lifting with varied muscle stiffness, planning trajectories and learning dynamics using via-points and the locally queried motor abundance library.

\addtolength{\textheight}{-12cm}   

\section*{ACKNOWLEDGMENT}
The author would like to thank Hiroaki Masuda for his mechanical maintenance of the upper limb robot, and Dr. Matthias Rolf for his suggestions in implementing directed online goal babbling.


\end{document}